\documentclass[11pt,a4paper]{article}
\PassOptionsToPackage{hyperindex,breaklinks}{hyperref} 
\PassOptionsToPackage{hyphens}{url}
\usepackage[hyperref]{naaclhlt2019}
\usepackage{times}
\usepackage{latexsym}
\usepackage[textsize=tiny,color=yellow]{todonotes}
\usepackage{booktabs}
\usepackage[normalem]{ulem}

\aclfinalcopy 

\title{Sentence Encoders on STILTs:\\ Supplementary Training on Intermediate Labeled-data Tasks}

\author{Jason Phang$^{1,}$\thanks{~~Equal contribution.} \\
  {\tt jasonphang@nyu.edu} \\\And
  Thibault F\'evry$^{1,*}$ \\
  {\tt thibault.fevry@nyu.edu} \\\And
  Samuel R. Bowman$^{1,2,3}$ \\
  {\tt bowman@nyu.edu} \\
  \AND
$^{1}$\normalfont Center for Data Science\\New York University\\60 Fifth Avenue\\New York, NY 10011\And
$^{2}$\normalfont Dept. of Linguistics\\New York University\\10 Washington Place\\New York, NY 10003\And
$^{3}$\normalfont Dept. of Computer Science\\New York University\\60 Fifth Avenue\\New York, NY 10011
  } 
\date{}

\date{}

\begin{document}
\maketitle
\begin{abstract}

Pretraining sentence encoders with language modeling and related unsupervised tasks has recently been shown to be very effective for language understanding tasks.
By supplementing language model-style pretraining with further training on data-rich supervised tasks, such as natural language inference, we obtain additional performance improvements on the GLUE benchmark.
Applying supplementary training on BERT \citep{devlin2018bert}, we attain a GLUE score of 81.8---the state of the art\footnote{As of 02/24/2019.} and a 1.4 point improvement over BERT. We also observe reduced variance across random restarts in this setting. 
Our approach yields similar improvements when applied to ELMo \citep{N18-1202} and \citet{radford2018improving}'s model. In addition, the benefits of supplementary training are particularly pronounced in data-constrained regimes, as we show in experiments with artificially limited training data. 

\end{abstract}

\section{Introduction}

Recent work has shown mounting evidence that pretraining sentence encoder neural networks on unsupervised tasks like language modeling, and then fine-tuning them on individual target tasks, can yield significantly better target task performance than could be achieved using target task training data alone \citep{P18-1031,radford2018improving,devlin2018bert}. Large-scale unsupervised pretraining in works like these seems to produce sentence encoders with substantial knowledge of the target language (which, so far, is generally English). These works have shown that the  one-size-fits-all approach of fine-tuning a large pretrained model with a thin output layer for a given task can achieve results as good or better than carefully-designed task-specific models without such pretraining. 

However, it is not obvious that the model parameters obtained during unsupervised pretraining should be \textit{ideally suited} to supporting this kind of transfer learning. Especially when only a small amount of training data is available for the target task, fine-tuning experiments are potentially brittle, and rely on the pretrained encoder parameters to be reasonably close to an ideal setting for the target task. During target task training, the encoder must learn and adapt enough to be able to solve the target task---potentially involving a very different input distribution and output label space than was seen in pretraining---but it must also avoid overfitting or catastrophic forgetting of what was learned during pretraining. 

This work explores the possibility that the use of a second stage of pretraining with data-rich intermediate supervised tasks might mitigate this brittleness, improving both the robustness and effectiveness of the resulting target task model. We name this approach, which is meant to be combined with existing approaches to pretraining, \textit{Supplementary Training on Intermediate Labeled-data Tasks} (STILTs). 

Experiments with sentence encoders on STILTs take the following form: (i) A model is first trained on an unlabeled-data task like language modeling that can teach it to reason about the target language; (ii) The model is then further trained on an intermediate, labeled-data task for which ample data is available; (iii) The model is finally fine-tuned further on the target task and evaluated. Our experiments evaluate STILTs as a means of improving target task performance on the GLUE benchmark suite \citep{wang2018glue}---a collection of language understanding tasks drawn from the NLP literature. 

We apply STILTs to three separate pretrained sentence encoders: BERT \citep{devlin2018bert}, GPT \citep{radford2018improving}, and a variant of ELMo \citep{N18-1202}.
We follow \citeauthor{radford2018improving}~and \citeauthor{devlin2018bert}~in our basic mechanism for fine-tuning both for the intermediate and final tasks, and use the following four intermediate tasks: (i) the Multi-Genre NLI Corpus \citep[MNLI;][]{DBLP:journals/corr/WilliamsNB17}, (ii) the Stanford NLI Corpus \citep[SNLI;][]{bowman2015large}, (iii) the Quora Question Pairs\footnote{\href{https://data.quora.com/First-Quora-Dataset-Release-Question-Pairs}{https://data.quora.com/First-Quora-Dataset-Release-\\Question-Pairs}} (QQP) dataset, and (iv) a custom fake-sentence-detection task based on the BooksCorpus dataset \citep{zhu2015aligning} using a method adapted from \citet{warstadt2018neural}. The use of MNLI and SNLI is motivated by prior work on using natual language inference tasks to pretrain sentence encoders \citep{DBLP:conf/emnlp/ConneauKSBB17,subramanian2018large,bowman2019looking}. QQP has a similar format and dataset scale, while requiring a different notion of sentence similarity. The fake-sentence-detection task is motivated by \citeauthor{warstadt2018neural}'s analysis on CoLA and linguistic acceptability, and adapted for our experiments.
These four tasks are a sample of data-rich supervised tasks that we can use to demonstrate the benefits of STILTs, but they do not represent an exhaustive exploration of the space of promising intermediate tasks.

We show that using STILTs yields significant gains across most of the GLUE tasks, across all three sentence encoders we used, and claims the state of the art on the overall GLUE benchmark. In addition, for the 24-layer version of BERT, which can require multiple random restarts for good performance on target tasks with limited training data, we find that STILTs substantially reduces the number of runs with degenerate results across random restarts. For instance, using STILTs with 5k training examples, we reduce the number of degenerate runs from five to one on SST and from two to none on STS. 

As we expect that any kind of pretraining will be most valuable in a limited training data regime, we also conduct a set of experiments where a model is fine-tuned on only 1k- or 5k-example subsamples of the target task training set. The results show that STILTs substantially improves model performance across most tasks in this downsampled data setting, even more so than in the full-data setting.

\section{Related Work}

In the area of pretraining for sentence encoders, \citet{zhang2018lessons} compare several pretraining tasks for syntactic target tasks, and find that language model pretraining reliably performs well.
\citet{peters2018dissecting} investigate the architectural choices behind ELMo-style pretraining with a fixed encoder, and find that the precise choice of encoder architecture strongly influences training speed, but has a relatively small impact on performance.
\citet{bowman2019looking} compare a variety of tasks for pretraining in an ELMo-style setting with no encoder fine-tuning. They conclude that language modeling generally works best among candidate single tasks for pretraining, but show some cases in which a cascade of a model pretrained on language modeling followed by another model pretrained on tasks like MNLI can work well. The paper introducing BERT \citep{devlin2018bert} briefly mentions encouraging results in a direction similar to ours: One footnote notes that unpublished experiments show ``substantial improvements on RTE from multitask training
with MNLI.''

Most prior work uses features from frozen, pretrained sentence encoders in downstream tasks. A more recent trend of fine-tuning the whole model for the target task from a pretrained state \citep{P18-1031, radford2018improving, devlin2018bert} has led to state-of-the-art results on several benchmarks. For that reason, we focus our analysis on the paradigm of fine-tuning the whole model for each task.

In the area of sentence-to-vector encoding, \citet{P18-1198} offer one of the most comprehensive suites of diagnostic tasks, and highlight the importance of ensuring that these models preserve lexical content information. 

In earlier work less closely tied to the unsupervised pretraining setup studied here,  \citet{bingel-sogaard:2017:EACLshort} and \citet{kerinec2018does} investigate the conditions under which task combinations can be productively combined in multitask learning. They show that multitask learning is more likely to work when the target task quickly plateaus and the auxiliary task keeps improving. They also report that gains are lowest when the Jensen-Shannon Divergence between the unigram distributions of tasks is highest, i.e when auxiliary and target tasks have different vocabulary.

In word representations, this work shares motivations with work on embedding space \textit{retrofitting} \citep{N15-1184} wherein a labeled dataset like WordNet is used to refine representations learned by an unsupervised embedding learning algorithm before those representations are used for a target task.

\begin{table*}[t]
\centering \small \setlength{\tabcolsep}{0.5em}
\begin{tabular}{lrrrrr@{/}rr@{/}rr@{/}rrrrr}
\toprule
\textbf{} & \multicolumn{1}{c}{\textbf{Avg}} & \multicolumn{1}{c}{\textbf{A.Ex}} & \multicolumn{1}{c}{\textbf{CoLA}} & \multicolumn{1}{c}{\textbf{SST}} & \multicolumn{2}{c}{\textbf{MRPC}} & \multicolumn{2}{c}{\textbf{QQP}} & \multicolumn{2}{c}{\textbf{STS}} & \multicolumn{1}{c}{\textbf{MNLI}} & \multicolumn{1}{c}{\textbf{QNLI}} & \multicolumn{1}{c}{\textbf{RTE}} & \multicolumn{1}{c}{\textbf{WNLI}} \\
\textbf{Training Set Size}&&&\multicolumn{1}{c}{\textit{8.5k}} & \multicolumn{1}{c}{\textit{67k}} & \multicolumn{2}{c}{\textit{3.7k}} & \multicolumn{2}{c}{\textit{364k}} & \multicolumn{2}{c}{\textit{7k}} & \multicolumn{1}{c}{\textit{393k}} & \multicolumn{1}{c}{\textit{108k}} & \multicolumn{1}{c}{\textit{2.5k}} & \multicolumn{1}{c}{\textit{634}}\\
\midrule
\midrule
\multicolumn{15}{c}{Development Set Scores} \\
\midrule
\textbf{BERT} & 80.8 & 78.4 & \textbf{62.1} & 92.5 & 89.0 & 92.3 & \textbf{91.5} & \textbf{88.5} & 90.3 & 90.1 & \textbf{86.2} & 89.4 & 70.0 & 56.3 \\
\textbf{BERT$\rightarrow$QQP} & 80.9 & 78.5 & 56.8 & 93.1 & 88.7 & 92.0 & \sout{91.5} & \sout{88.5} & 90.9 & 90.7 & 86.1 & 89.5 & 74.7 & 56.3 \\
\textbf{BERT$\rightarrow$MNLI} & 82.4 & 80.5 & 59.8 & \textbf{93.2} & \textbf{89.5} & \textbf{92.3} & 91.4 & 88.4 & \textbf{91.0} & \textbf{90.8} & \sout{86.2} & \textbf{90.5} & \textbf{83.4} & 56.3 \\
\textbf{BERT$\rightarrow$SNLI} & 81.4 & 79.2 & 57.0 & 92.7 & 88.5 & 91.7 & 91.4 & 88.4 & 90.7 & 90.6 & 86.1 & 89.8 & 80.1 & 56.3 \\
\textbf{BERT$\rightarrow$Real/Fake} & 77.4 & 74.3 & 52.4 & 92.1 & 82.8 & 88.5 & 90.8 & 87.5 & 88.7 & 88.6 & 84.5 & 88.0 & 59.6 & 56.3 \\
\textbf{BERT, Best of Each} & \textbf{82.6} & \textbf{80.8} & \textbf{62.1} & \textbf{93.2} & \textbf{89.5} & \textbf{92.3} & \textbf{91.5} & \textbf{88.5} & \textbf{91.0} & \textbf{90.8} & \textbf{86.2} & \textbf{90.5} & \textbf{83.4} & 56.3 \\
\midrule
\textbf{GPT} & 75.4 & 72.4 & \textbf{50.2} & \textbf{93.2} & 80.1 & 85.9 & 89.4 & 85.9 & 86.4 & 86.5 & \textbf{81.2} & 82.4 & 58.1 & 56.3 \\
\textbf{GPT$\rightarrow$QQP} & 76.0 & 73.1 & 48.3 & 93.1 & 83.1 & 88.0 & \sout{89.4} & \sout{85.9} & 87.0 & 86.9 & 80.7 & 82.6 & 62.8 & 56.3 \\
\textbf{GPT$\rightarrow$MNLI} & 76.7 & 74.2 & 45.7 & 92.2 & \textbf{87.3} & \textbf{90.8} & 89.2 & 85.3 & 88.1 & 88.0 & \sout{81.2} & \textbf{82.6} & \textbf{67.9} & 56.3 \\
\textbf{GPT$\rightarrow$SNLI} & 76.0 & 73.1 & 41.5 & 91.9 & 86.0 & 89.9 & 89.9 & 86.6 & \textbf{88.7} & \textbf{88.6} & 81.1 & 82.2 & 65.7 & 56.3 \\
\textbf{GPT$\rightarrow$Real/Fake} & 76.6 & 73.9 & 49.5 & 91.4 & 83.6 & 88.6 & \textbf{90.1} & \textbf{86.9} & 87.9 & 87.8 & 81.0 & 82.5 & 66.1 & 56.3 \\
\textbf{GPT, Best of Each} & \textbf{77.5} & \textbf{75.9} & \textbf{50.2} & \textbf{93.2} & \textbf{87.3} & \textbf{90.8} & \textbf{90.1} & \textbf{86.9} & \textbf{88.7} & \textbf{88.6} & \textbf{81.2} & \textbf{82.6} & \textbf{67.9} & 56.3 \\
\midrule
\textbf{ELMo} & 63.8 & 59.4 & 15.6 & 84.9 & 69.9 & 80.6 & 86.4 & 82.2 & 64.5 & 64.4 & 69.4 & 73.0 & 50.9 & 56.3 \\
\textbf{ELMo$\rightarrow$QQP} & 64.8 & 61.7 & 16.6 & 87.0 & 73.5 & 82.4 & \sout{86.4} & \sout{82.2} & 71.6 & 72.0 & 63.9 & 73.4 & 52.0 & 56.3 \\
\textbf{ELMo$\rightarrow$MNLI} & 66.4 & 62.8 & 16.4 & 87.6 & 73.5 & 83.0 & 87.2 & 83.1 & \textbf{75.2} & \textbf{75.8} & \sout{69.4} & 72.4 & \textbf{56.3} & 56.3 \\
\textbf{ELMo$\rightarrow$SNLI} & 66.4 & 62.7 & 14.8 & \textbf{88.4} & \textbf{74.0} & \textbf{82.5} & \textbf{87.3} & \textbf{83.1} & 74.1 & 75.0 & 69.7 & \textbf{74.0} & 56.0 & 56.3 \\
\textbf{ELMo$\rightarrow$Real/Fake} & 66.9 & 63.3 & \textbf{27.3} & 87.8 & 72.3 & 81.3 & 87.1 & 83.1 & 70.3 & 70.6 & \textbf{70.3} & 73.7 & 54.5 & 56.3 \\
\textbf{ELMo, Best of Each} & \textbf{68.0} & \textbf{64.8} & \textbf{27.3} & \textbf{88.4} & \textbf{74.0} & \textbf{82.5} & \textbf{87.3} & \textbf{83.1} & \textbf{75.2} & \textbf{75.8} & \textbf{70.3} & \textbf{74.0} & \textbf{56.3} & 56.3 \\
\midrule
\multicolumn{15}{c}{Test Set Scores} \\
\midrule
\textbf{BERT} & 80.4 & 79.4 & 60.5 & \textbf{94.9} & 85.4 & 89.3 & \textbf{89.3} & \textbf{72.1} & 87.6 & 86.5 & \textbf{86.3} & \textbf{91.1} & 70.1 & 65.1 \\
\textbf{BERT on STILTs} & \textbf{81.8} & \textbf{81.4} & \textbf{62.1} & 94.3 & \textbf{89.8} & \textbf{86.7} & 89.4  & 71.9 & \textbf{88.7} & \textbf{88.3} & 86.0 & \textbf{91.1} & \textbf{80.1} & 65.1 \\
\midrule
\textbf{GPT} & 74.1 & 71.9 & 45.4 & 91.3 & 82.3 & 75.7 & \textbf{88.5} & \textbf{70.3} & 82.0 & 80.0 & \textbf{81.8} & 88.1 & 56.0 & 65.1 \\
\textbf{GPT on STILTs} & \textbf{76.9} & \textbf{75.9} & \textbf{47.2} & \textbf{93.1} & \textbf{87.7} & \textbf{83.7} & 88.1 & 70.1 & \textbf{85.3} & \textbf{84.8} & 80.7 & \textbf{87.2} & \textbf{69.1} & 65.1 \\
\midrule
\textbf{ELMo} & 62.2 & 59.0 & 16.2 & \textbf{87.1} & 79.7 & 69.1 & 84.9 & 63.9 & 64.3 & 63.9 & 69.0 & 57.1 & 52.3 & 65.1 \\
\textbf{ELMo on STILTs} & \textbf{65.9} & \textbf{63.8} & \textbf{30.3} & 86.5 & \textbf{82.0} & \textbf{73.9} & \textbf{85.2} & \textbf{64.4} & \textbf{71.8} & \textbf{71.4} & \textbf{69.7} & \textbf{62.6} & \textbf{54.4} & 65.1 \\
\bottomrule
\end{tabular}
\caption{GLUE results with and without STILTs, fine-tuning on full training data of each target task. \textbf{Bold} marks the best within each section. \sout{Strikethrough} indicates cases where the intermediate task is the same as the target task---we substitute the baseline result for that cell. \textit{A.Ex} is the average excluding MNLI and QQP because of the overlap with intermediate tasks. See text for discussion of WNLI results. Test results \textit{on STILTs} uses the supplementary training regime for each task based on the performance on the development set, corresponding to the numbers shown in \textit{Best of Each}. 
The aggregated GLUE scores differ from the public leaderboard because we report performance on QNLIv1.
}
\label{tab:glueresults}
\end{table*}

\section{Methods}

\paragraph{Pretrained Sentence Encoders}

We primarily study the impact of STILTs on three sentence encoders: BERT \citep{devlin2018bert}, GPT \citep{radford2018improving} and ELMo \citep{N18-1202}.  These models are distributed with pretrained weights from their respective authors, and are the best performing sentence encoders as measured by GLUE benchmark performance at time of writing. All three models are pretrained with large amounts of unlabeled text. ELMo uses a BiLSTM architecture whereas BERT and GPT use the Transformer architecture \citep{vaswani2017attention}. These models are also trained with different objectives and corpora.
BERT is a bi-directional Transformer trained on BooksCorpus \citep{bookcorpus} and English Wikipedia, with a masked-language model and next sentence prediction objective. GPT is uni-directional masked Transformer trained only on BooksCorpus with a standard language modeling objective. 
ELMo is trained on the 1B Word Benchmark \citep{billionword} with a standard language modeling objective. 

For all three pretrained models, we follow BERT and GPT in using an inductive approach to transfer learning, in which the model parameters learned during pretraining are used to initialize a target task model, but are not fixed and do not constrain the solution learned for the target task. This stands in contrast to the approach originally used for ELMo \citep{peters2018dissecting} and for earlier methods like \citet{mccann2017learned} and \citet{subramanian2018large}, in which a sentence encoder component is pretrained and then attached to a target task model as a non-trainable input layer.

To implement intermediate-task and target-task training for GPT and ELMo, we use the public \texttt{jiant} transfer learning toolkit,\footnote{\href{https://github.com/jsalt18-sentence-repl/jiant}{https://github.com/jsalt18-sentence-repl/jiant}} which is built on AllenNLP \citep{Gardner2017AllenNLP} and PyTorch \citep{paszke2017automatic}. 
For BERT, we use the publicly available implementation of BERT released by \citet{devlin2018bert}, ported into PyTorch\citep{paszke2017automatic} by HuggingFace\footnote{\href{https://github.com/huggingface/pytorch-pretrained-BERT}{https://github.com/huggingface/pytorch-pretrained-BERT}}.

\paragraph{Target Tasks and Evaluation}
We evaluate on the nine target tasks 
in the GLUE benchmark \citep{wang2018glue}. These include MNLI, QQP, and seven others: acceptability classification with CoLA \citep{warstadt2018neural}; binary sentiment classification with SST \citep{socher2013recursive}; semantic similarity with the MSR Paraphrase Corpus \citep[MRPC;][]{dolan2005automatically} and STS-Benchmark \citep[STS;][]{cer2017semeval}; and textual entailment with a subset of the RTE challenge corpora \citep[][et seq.]{dagan2006pascal}, and data from SQuAD \citep[QNLI,][]{rajpurkar2016squad}\footnote{A newer version of QNLI was recently released by the maintainers of GLUE benchmark. All reported numbers in this work, including the aggregated GLUE score, reflect evaluation on the older version of QNLI (QNLIv1).} and the Winograd Schema Challenge \citep[WNLI,][]{levesque2011winograd} converted to entailment format as in \citet{white2017inference}. Because of the adversarial nature of WNLI, our models do not generally perform better than chance, and we follow the recipe of \citet{devlin2018bert} by predicting the most frequent label for all examples. 

\begin{table*}[t]
\centering \small \setlength{\tabcolsep}{0.5em}
\begin{tabular}{lrrrrr@{/}rr@{/}rr@{/}rrrrr}
\toprule
\textbf{} & \multicolumn{1}{c}{\textbf{Avg}} & \multicolumn{1}{c}{\textbf{A.Ex}} & \multicolumn{1}{c}{\textbf{CoLA}} & \multicolumn{1}{c}{\textbf{SST}} & \multicolumn{2}{c}{\textbf{MRPC}} & \multicolumn{2}{c}{\textbf{QQP}} & \multicolumn{2}{c}{\textbf{STS}} & \multicolumn{1}{c}{\textbf{MNLI}} & \multicolumn{1}{c}{\textbf{QNLI}} & \multicolumn{1}{c}{\textbf{RTE}} & \multicolumn{1}{c}{\textbf{WNLI}} \\
\textbf{Training Set Size}&&&\multicolumn{1}{c}{\textit{8.5k}} & \multicolumn{1}{c}{\textit{67k}} & \multicolumn{2}{c}{\textit{3.7k}} & \multicolumn{2}{c}{\textit{364k}} & \multicolumn{2}{c}{\textit{7k}} & \multicolumn{1}{c}{\textit{393k}} & \multicolumn{1}{c}{\textit{108k}} & \multicolumn{1}{c}{\textit{2.5k}} & \multicolumn{1}{c}{\textit{634}}\\
\midrule
\midrule
\multicolumn{14}{c}{At Most 5k Training Examples for Target Tasks} \\
\midrule
\textbf{BERT} & 78.3 & 78.1 & \textbf{60.6} & \textbf{93.5} & 87.3 & 91.0 & 83.1 & 78.6 & 90.2 & 89.8 & 77.1 & 82.8 & 74.0 & 56.3 \\
\textbf{BERT$\rightarrow$QQP} & 77.6 & 77.3 & 55.3 & 92.0 & 88.0 & 91.4 & \sout{83.1} & \sout{78.6} & 90.7 & 90.5 & 75.9 & 81.6 & 76.5 & 56.3 \\
\textbf{BERT$\rightarrow$MNLI} & 79.5 & 79.7 & 59.6 & 92.4 & \textbf{89.5} & \textbf{92.5} & \textbf{83.7} & \textbf{78.1} & \textbf{91.1} & \textbf{90.6} & \sout{77.1} & \textbf{83.9} & \textbf{83.4} & 56.3 \\
\textbf{BERT$\rightarrow$SNLI} & 78.8 & 78.2 & 56.6 & 91.5 & 88.2 & 91.6 & 83.0 & 77.9 & 90.8 & 90.6 & \textbf{80.6} & 82.7 & 80.5 & 56.3 \\

\textbf{BERT$\rightarrow$Real/Fake} & 71.0 & 71.7 & 53.6 & 88.9 & 82.6 & 87.6 & 81.7 & 76.1 & 88.4 & 88.4 & 59.1 & 74.1 & 54.9 & 56.3 \\
\textbf{BERT, Best of Each} & \textbf{80.1} & \textbf{79.9} & \textbf{60.6} & \textbf{93.5} & \textbf{89.5} & \textbf{92.5} & \textbf{83.7} & \textbf{78.1} & \textbf{91.1} & \textbf{90.6} & \textbf{80.6} & \textbf{83.9} & \textbf{83.4} & 56.3 \\
\midrule
\textbf{GPT} & 71.6 & 71.2 & \textbf{50.8} & \textbf{91.1} & 81.4 & 87.1 & 79.5 & 73.8 & 87.6 & 87.4 & 68.8 & 73.1 & 56.3 & 56.3 \\
\textbf{GPT$\rightarrow$QQP} & 65.2 & 63.3 & 0.0 & 82.0 & 82.8 & 87.7 & \sout{79.5} & \sout{73.8} & 87.4 & 87.3 & 65.1 & 71.6 & 62.8 & 56.3 \\
\textbf{GPT$\rightarrow$MNLI} & 72.3 & 71.8 & 35.3 & 89.4 & \textbf{86.8} & \textbf{90.8} & \textbf{81.6} & \textbf{76.3} & 88.8 & 88.7 & \sout{68.8} & 74.1 & \textbf{70.4} & 56.3 \\
\textbf{GPT$\rightarrow$SNLI} & 72.3 & 70.2 & 29.6 & 89.2 & 86.3 & 90.2 & 81.6 & 76.0 & \textbf{89.5} & \textbf{89.4} & \textbf{78.3} & \textbf{74.7} & 66.4 & 56.3 \\
\textbf{GPT$\rightarrow$Real/Fake} & 71.4 & 69.3 & 45.1 & 87.8 & 78.2 & 85.2 & 80.6 & 75.4 & 87.8 & 87.5 & 77.5 & 72.2 & 56.3 & 56.3 \\
\textbf{GPT, Best of Each} & \textbf{75.4} & \textbf{74.3} & \textbf{50.8} & \textbf{91.1} & \textbf{86.8} & \textbf{90.8} & \textbf{81.6} & \textbf{76.3} & \textbf{89.5} & \textbf{89.4} & \textbf{78.3} & \textbf{74.7} & \textbf{70.4} & 56.3 \\
\midrule
\midrule
\multicolumn{14}{c}{At Most 1k Training Examples for Target Tasks} \\
\midrule
\textbf{BERT} & 74.2 & 74.5 & \textbf{54.0} & \textbf{91.1} & 83.8 & 88.4 & 79.9 & 73.8 & 88.1 & 87.9 & 69.7 & 77.0 & 69.0 & 56.3 \\
\textbf{BERT$\rightarrow$QQP} & 73.2 & 73.5 & 47.5 & 89.7 & 82.1 & 86.9 & \sout{79.9} & \sout{73.8} & 88.6 & 88.5 & 67.5 & 76.4 & 71.5 & 56.3 \\
\textbf{BERT$\rightarrow$MNLI} & 75.1 & 75.6 & 44.0 & 90.5 & \textbf{85.5} & \textbf{90.0} & 80.3 & 74.3 & \textbf{88.7} & \textbf{88.7} & \sout{69.7} & \textbf{79.0} & \textbf{82.7} & 56.3 \\
\textbf{BERT$\rightarrow$SNLI} & 75.5 & 74.7 & 47.6 & 89.3 & 82.8 & 87.8 & \textbf{80.6} & \textbf{74.1} & 87.8 & 88.1 & \textbf{78.6} & 77.6 & 79.1 & 56.3 \\
\textbf{BERT$\rightarrow$Real/Fake} & 63.9 & 67.5 & 43.9 & 72.5 & 78.9 & 84.7 & 74.1 & 68.4 & 82.4 & 83.2 & 35.3 & 69.7 & 61.7 & 56.3 \\
\textbf{BERT, Best of Each} & \textbf{77.3} & \textbf{77.1} & \textbf{54.0} & \textbf{91.1} & \textbf{85.5} & \textbf{90.0} & \textbf{80.6} & \textbf{74.1} & \textbf{88.7} & \textbf{88.7} & \textbf{78.6} & \textbf{79.0} & \textbf{82.7} & 56.3 \\
\midrule
\textbf{GPT} & 64.5 & 64.8 & 33.4 & 85.3 & 70.1 & 81.3 & 75.3 & 67.7 & 80.8 & 80.8 & 55.7 & 66.7 & 54.9 & 56.3 \\
\textbf{GPT$\rightarrow$QQP} & 64.6 & 64.6 & 23.0 & \textbf{87.0} & 74.8 & 83.2 & \sout{75.3} & \sout{67.7} & 84.4 & 84.3 & 57.8 & 67.1 & 55.2 & 56.3 \\
\textbf{GPT$\rightarrow$MNLI} & 65.2 & 65.2 & 13.3 & 86.2 & 79.2 & 85.8 & \textbf{78.4} & \textbf{70.5} & 86.2 & 86.1 & \sout{55.7} & 68.6 & \textbf{63.2} & 56.3 \\
\textbf{GPT$\rightarrow$SNLI} & 67.5 & 64.9 & 13.4 & 85.7 & \textbf{80.1} & \textbf{86.2} & 77.2 & 70.0 & \textbf{87.5} & \textbf{87.5} & \textbf{76.8} & \textbf{70.3} & 60.6 & 56.3 \\
\textbf{GPT$\rightarrow$Real/Fake} & 65.3 & 62.5 & \textbf{36.3} & 69.7 & 69.6 & 79.6 & 75.5 & 69.4 & 84.7 & 84.8 & 74.6 & 69.1 & 50.2 & 56.3 \\
\textbf{GPT, Best of Each} & \textbf{70.6} & \textbf{68.9} & \textbf{36.3} & \textbf{87.0} & \textbf{80.1} & \textbf{86.2} & \textbf{78.4} & \textbf{70.5} & \textbf{87.5} & \textbf{87.5} & \textbf{76.8} & \textbf{70.3} & \textbf{63.2} & 56.3 \\
\bottomrule
\end{tabular}
\caption{Results on the GLUE development set based on fine-tuning on only a subset of target-task data, simulating data scarce scenarios. \textbf{Bold} indicates the best within each section. \sout{Strikethrough} indicates cases where the intermediate task is the same as the target task: We substitute the baseline result for that cell. \textit{A.Ex} is the average excluding MNLI and QQP, because of their overlap with the candidate intermediate tasks. See text for discussion of WNLI results.}
\label{tab:glueresults_small}
\end{table*}

\begin{figure*}[t]
    \makebox[\linewidth]{
      \includegraphics[keepaspectratio=true,scale=0.5]{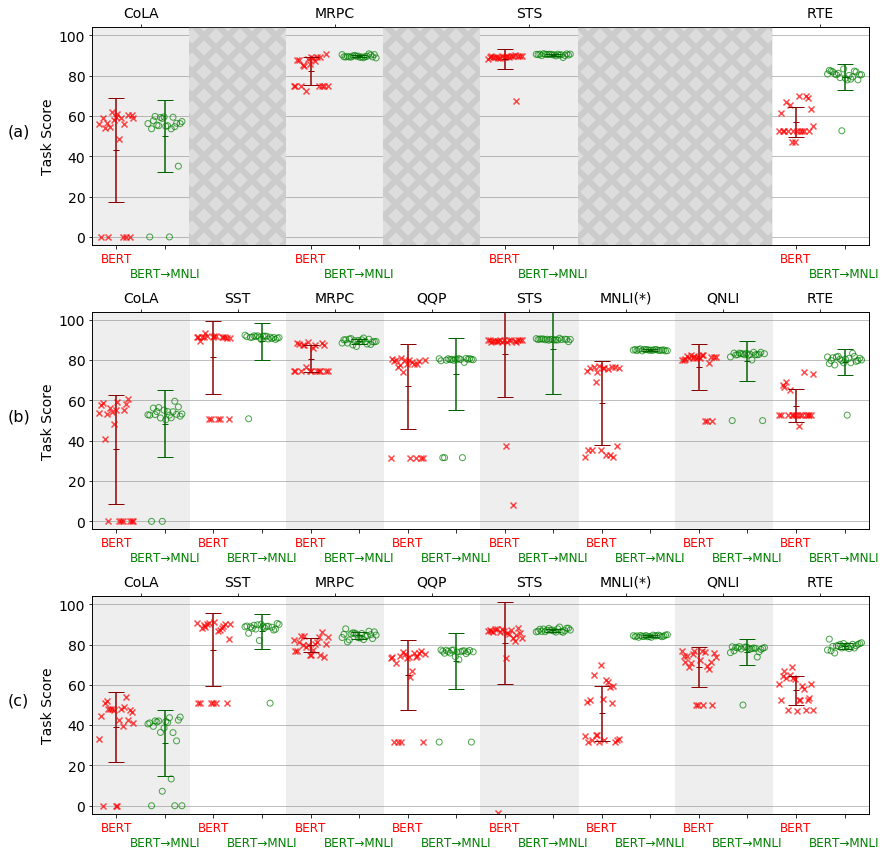}}
    \captionof{figure}{Distribution of task scores across 20 random restarts for BERT, and BERT with intermediary fine-tuning on MNLI. Each cross represents a single run. Error lines show mean$\pm$1std. (a) Fine-tuned on all data, for tasks with $<$10k training examples. (b) Fine-tuned on no more than 5k examples for each task. (c) Fine-tuned on no more than 1k examples for each task. (*) indicates that the intermediate task is the same as the target task.}
    \label{fig:bertboxplot}
\end{figure*}

Most of our experiments---including all of our experiments using downsampled training sets for our target tasks---are evaluated on the \textit{development set} of GLUE. Based on the results on the development set, we choose the best intermediate-task training scheme for each task and submit the best-per-task model for evaluation on the test set on the public leaderboard.

\paragraph{Intermediate Task Training}
Our experiments follow the standard pretrain-then-fine-tune approach, except that we add a supplementary training phase on an intermediate task before target-task fine-tuning. We call this approach \textit{BERT on STILTs}, \textit{GPT on STILTs} and \textit{ELMo on STILTS} for the respective models. We evaluate a sample of four intermediate tasks, which were chosen to represent readily available data-rich sentence-level tasks similar to those in GLUE: 
(i) textual entailment with MNLI; (ii) textual entailment with SNLI; (iii) paraphrase detection with QQP; and (iv) a custom fake-sentence-detection task. 

Our use of MNLI is motivated by prior successes with MNLI pretraining by \citet{P18-1198} and \citet{subramanian2018large}.
We include the single-genre captions-based SNLI in addition to the multi-genre MNLI to disambiguate between the benefits of domain shift and task shift from supplementary training on natural language inference. QQP is included as we believed it could improve performance on sentence similarity tasks such as MRPC and STS. Lastly, we construct a fake-sentence-detection task based on the BooksCorpus dataset in the style of \citeauthor{warstadt2018neural}. Importantly, because both GPT and BERT are pretrained on BooksCorpus, the fake-sentence-detection enables us to isolate the impact of task shift from domain shift from the pretaining corpus. We construct this task by sampling sentences from BooksCorpus, and fake sentences are generated by randomly swapping 2--4 pairs of words in the sentence. We generate a dataset of 600,000 sentences with a 50/50 real/fake split for this intermediate task. 

\paragraph{Training Details}

Unless otherwise stated, for replications and both stages of our STILTs experiments, we follow the model formulation and training regime of BERT and the GPT specified in \citeauthor{devlin2018bert}~and \cite{radford2018improving}~respectively. Specifically, for both models we use a three-epoch training limit for both supplementary training and target-task fine-tuning. We use a fresh optimizer for each phase of training. For each task, we add only a single task-specific, randomly initialized output layer to the pretrained Transformer model, following the setup laid out by each respective work. For our baseline, we do not fine-tune on any intermediate task: Other than the batch size, this is equivalent to the formulation presented in the papers introducing BERT and GPT respectively and serves as our attempt to replicate their results.

For BERT, we use a batch size of 24 and a learning rate of 2e-5. This is within the range of hyperparameters recommended by the authors and initial experiments showed promising results. We use the larger, 24-layer version of BERT, which is the state of the art on the GLUE benchmark. For this model, fine-tuning can be unstable on small data sets---hence, for the tasks with limited data (CoLA, MRPC, STS, RTE), we perform 20 random restarts for each experiment and report the results of the model that performed best on the validation set. 

For GPT, we choose the largest batch size out of 8/16/32 that a single GPU can accommodate. We use the version with an auxiliary language modeling objective in fine-tuning, corresponding to the entry on the GLUE leaderboard.\footnote{\citet{radford2018improving} introduced two versions of GPT: one which includes an auxiliary language modeling objective when fine-tuning, and one without.} 

For ELMo, to facilitate a fair comparison with GPT and BERT, we adopt a similar fine-tuning setup where all the weights are fine-tuned. This differs from the original ELMo setup that freezes ELMo weights and trains an additional encoder module when fine-tuning. The details of our ELMo setup are described in Appendix~\ref{sec:elmodetails}.

We also run our main experiment on the 12-layer BERT and the non-LM fine-tuned GPT. These results are in Table~\ref{tab:glueresults_nolm} in the Appendix. 

\paragraph{Multitask Learning Strategies}
To compare STILTs to alternative multitask learning regimes, we also experiment with the following two approaches: (i) a single phase of fine-tuning simultaneously on both a intermediate task and the target task (ii) fine-tuning simultaneously on a intermediate task and the target task, and then doing an additional phase of fine-tuning on the target task only. In the multitask learning phase, for both approaches, training steps are sampled proportionally to the sizes of the respective training sets and we do not weight the losses. 
 
\paragraph{Models and Code} 
Our pretrained models and code for BERT on STILTs can be found at \href{https://github.com/zphang/pytorch-pretrained-BERT}{https://github.com/zphang/pytorch-pretrained-BERT}, which is a fork of the Hugging Face implementation. We used the \texttt{jiant} framework experiments on GPT and ELMo.
 
\section{Results}

Table~\ref{tab:glueresults} shows our results on GLUE with and without STILTs. 
Our addition of supplementary training boosts performance across many of the two-sentence tasks. We also find that most of the gains are on tasks with limited data. On each of our STILTs models, we show improved overall GLUE scores on the development set. Improvements from STILTs tend to be larger for ELMo and GPT and somewhat smaller for BERT. On the other hand, for pairs of pretraining and target tasks that are close, such as MNLI and RTE, we indeed find a marked improvement in performance from STILTs. For the two single-sentence tasks---the syntax-oriented CoLA task and the SST sentiment task---we find somewhat deteriorated performance. For CoLA, this mirrors results reported in \citet{bowman2019looking}, who show that few pretraining tasks other than language modeling offer any advantage for CoLA. The \textit{Best of Each} score is computed based on taking the best score for each task, including no STILTs. 

On the test set, we see similar performance gains across most tasks. Here, we compute the results for each model \textit{on STILTs}, which shows scores from choosing the best corresponding model based on development set scores and evaluating on the test set. These also correspond to the selected models for \textit{Best of Each} above.\footnote{For BERT, we run an additional 80 random restarts--100 random restarts in total--for the tasks with limited data, and select the best model based on validation score for test evaluation} For both BERT and GPT, we show that using STILTs leads to improvements in test set performance
improving on the reported baseline by 1.4 points and setting the state of the art for the GLUE benchmark, while GPT on STILTs achieves a score of 76.9, improving on the baseline by 2.8 points, and significantly closing the gap between GPT and the 12-layer BERT model with a similar number of parameters, which attains a GLUE score of 78.3.


\paragraph{Limited Target-Task Data}

Table~\ref{tab:glueresults_small} shows the same models fine-tuned on 5k training examples and 1k examples for each task, selected randomly without replacement. Artificially limiting the size of the training set allows us to examine the effect of STILTs in data constrained contexts. For tasks with training sets that are already smaller than these limits, we use the training sets as-is. For BERT, we show the maximum task performance across 20 random restarts for all experiments, and the data subsampling is also random for each restart. 

The results show that the benefits of supplementary training are generally more pronounced in these settings, with performance in several tasks showing improvements of more than 10 points. CoLA and SST are again the exceptions: Both tasks deteriorated moderately with supplementary training, and CoLA trained with the auxiliary language modeling objective in particular showed highly unstable results when trained on small amounts of data.

We see one obvious area for potential improvement: In our experiments, we follow the recipe for fine-tuning from the original works as closely as possible, only doing supplementary training and fine-tuning for three epochs each. Particularly in the case of the artificially data-constrained tasks, we expect that performance could be improved with more careful tuning of the training duration and learning rate schedule.


\paragraph{Fine-Tuning Stability}
In the work that introduced BERT, \citeauthor{devlin2018bert}~highlight that the larger, 24-layer version of BERT is particularly prone to degenerate performance on tasks with small training sets, and that multiple random restarts may be required to obtain a usable model. In Figure~\ref{fig:bertboxplot}, we plot the distribution of performance scores for 20 random restarts for each task, using all training data and maximum of 5k or 1k training examples. For conciseness, we only show results for BERT without STILTs, and BERT with intermediate fine-tuning on MNLI. We omit the random restarts for tasks with training sets of more than 10k examples, consistent with our training methodology. 

We show that, in addition to improved performance, using STILTs significantly reduces the variance of performance across random restarts. A large part of reduction can be attributed to the far fewer number of degenerate runs---performance outliers that are close to random guessing. This effect is consistent across target tasks, though the magnitude varies from task to task. For instance, although we show above that STILTs with our four intermediate tasks does not improve model performance in CoLA and SST, using STILTs nevertheless reduces the variance across runs as well as the number of degenerate fine-tuning results. 

\paragraph{Multitask Learning and STILTs} We investigate whether setups that leverage multitask learning are more effective than STILTs. We highlight results from one of the cases with the largest improvement: GPT with intermediary fine-tuning on MNLI with RTE as the target task. To better isolate the impact of multitask learning, we exclude the auxiliary language modeling training objective in this experiment. Table~\ref{tab:mtexp} shows all setups improve compared to only fine-tuning, with the STILTs format of consecutive single-task fine-tuning having the largest improvement. Although this does not represent an in-depth inquiry of all the ways to leverage multitask learning and balance multiple training objective, naive multitask learning appears to yield worse performance than STILTs, at potentially greater computational cost.

\begin{table}[t]
\centering \small \setlength{\tabcolsep}{0.5em}
\begin{tabular}{lr}
\toprule
\textbf{Model} & \textbf{RTE accuracy}\\
\midrule
\textbf{GPT $\rightarrow$ RTE} &  54.2\\
\textbf{GPT $\rightarrow$ MNLI $\rightarrow$ RTE} & \textbf{70.4}\\
\textbf{GPT $\rightarrow$ \{MNLI, RTE\}} & 68.6 \\
\textbf{GPT $\rightarrow$ \{MNLI, RTE\} $\rightarrow$ RTE} & 67.5 \\
\bottomrule
\end{tabular}
\caption{Comparison of STILTs against multitask learning setups for GPT, with MNLI as the intermediate task, and RTE as the target task. GPT is fine-tuned without the auxiliary language modeling objective in this experiment. Both intermediary and final fine-tuning task(s) are delineated here, in contrast to Table~\ref{tab:glueresults} and Table~\ref{tab:glueresults_small} where we omit the name of the target-task.}
\label{tab:mtexp}
\end{table}

\section{Discussion}

Broadly, we have shown that, across three different sentence encoders with different architectures and pretraining schemes, STILTs can leads to performance gains on many downstream target tasks. However, this benefit is not uniform. We find that sentence pair tasks seem to benefit more from supplementary training than single-sentence ones. We also find that tasks with little training data benefit much more from supplementary training. Indeed, when applied to RTE, supplementary training on the related MNLI task leads to a eight-point increase in test set score for BERT. 

Overall, the benefit of STILTs is smaller for BERT than for GPT and ELMo. One possible reason is that BERT is better conditioned for fine-tuning for classification tasks, such as those in the GLUE Benchmark. Indeed, GPT uses the hidden state corresponding to the last token of the sentence as a proxy to encode the whole sentence, but this token is not used for classification during pre-training. On the other hand, BERT has a $<$CLS$>$ token which is used for classification during pre-training for their additional next-sentence-prediction objective. This token is then used in fine-tuning for classification. When adding STILTs to GPT, we bridge that gap by training the last token with the classification objective of the intermediary task. This might explain why fake-sentence-detection is a broadly beneficial task for GPT and not for BERT: Since fake-sentence-detection uses the same corpus that GPT and BERT are pretrained on, it is likely that the improvements we find for GPT are due to the better conditioning of this sentence-encoding token.

Applying STILTs also comes with little complexity or computational overhead. The same infrastructure used to fine-tune BERT or GPT models can be used to perform supplementary training. The computational cost of the supplementary training phase is another phase of fine-tuning, which is small compared to the cost of training the original model. In addition, in the case of BERT, the smaller number of degenerate runs induced by STILTs will reduce the computational cost of a full training procedure in some settings.

Our results also show where STILTs may be ineffective or counterproductive. In particular, we show that most of our intermediate tasks were actually detrimental to the single-sentence tasks in GLUE. The interaction between the intermediate task, the target task, and the use of the auxiliary language modeling objective is a subject due for further investigation. Moreover, the four intermediary training tasks we chose represent only a small sample of potential tasks, and it is likely that a more expansive survey might yield better performance on different downstream tasks. Therefore, for best target task performance, we recommend experimenting with supplementary training with several closely-related data-rich tasks and use the development set to select the most promising approach for each task, as in the \textit{Best of Each} formulation shown in Table~\ref{tab:glueresults}.

\section{Conclusion}

This work represents only an initial investigation into the benefits of supplementary supervised pretraining. More work remains to be done to firmly establish when methods like STILTs can be productively applied and what criteria can be used to predict which combinations of intermediate and target tasks should work well. Nevertheless, in our initial work with four example intermediate training tasks, we showed significant gains from applying STILTs to three sentence encoders, BERT, GPT and ELMo, and set the state of the art on the GLUE benchmark with BERT on STILTs. STILTs also helps to significantly stabilize training in unstable training contexts, such as when using BERT on tasks with little data. Finally, we show that in data-constrained regimes, the benefits of using STILTs are even more pronounced, yielding up to 10 point score improvements on some intermediate/target task pairs.

\section*{Acknowledgments}
We would like to thank Alex Wang, Ilya Kulikov, Nikita Nangia and Phu Mon Htut for their helpful feedback.

\bibliography{naaclhlt2019}
\bibliographystyle{acl_natbib}

\newpage
\clearpage
\appendix

\section{ELMo on STILTs}
\label{sec:elmodetails}

\paragraph{Experiment setup}
We use the same architecture as \citet{N18-1202} for the non-task-specific parameters. For task-specific parameters, we use the layer weights and the task weights described in the paper, as well as a classifier composed of max-pooling with projection and a logistic regression classifier. In contrast to the GLUE baselines and to \citet{bowman2019looking}, we refrain from adding many non-LM pretrained parameters by not using pair attention nor an additional encoding layer. The whole model, including ELMo parameters, is trained during both supplementary training on the intermediate task and target-task tuning. For two-sentence tasks, we follow the model design of \citet{wang2018glue} rather than that of \citet{radford2018improving}, since early experiments showed better performance with the former. Consequently, we run the shared encoder on the two sentences $u$ and $u'$ independently and then use $[u' ; v'; | u' - v' |; u'*v']$ for our task-specific classifier. We use the default optimizer and learning rate schedule from \texttt{jiant}.

\begin{table*}[t]
\centering \small \setlength{\tabcolsep}{0.5em}
\begin{tabular}{lrrrrr@{/}rr@{/}rr@{/}rrrrr}
\toprule
\textbf{} & \multicolumn{1}{c}{\textbf{Avg}} & \multicolumn{1}{c}{\textbf{AvgEx}} & \multicolumn{1}{c}{\textbf{CoLA}} & \multicolumn{1}{c}{\textbf{SST}} & \multicolumn{2}{c}{\textbf{MRPC}} & \multicolumn{2}{c}{\textbf{QQP}} & \multicolumn{2}{c}{\textbf{STS}} & \multicolumn{1}{c}{\textbf{MNLI}} & \multicolumn{1}{c}{\textbf{QNLI}} & \multicolumn{1}{c}{\textbf{RTE}} & \multicolumn{1}{c}{\textbf{WNLI}} \\
\textbf{Training Set Size}&&&\multicolumn{1}{c}{\textit{8.5k}} & \multicolumn{1}{c}{\textit{67k}} & \multicolumn{2}{c}{\textit{3.7k}} & \multicolumn{2}{c}{\textit{364k}} & \multicolumn{2}{c}{\textit{7k}} & \multicolumn{1}{c}{\textit{393k}} & \multicolumn{1}{c}{\textit{108k}} & \multicolumn{1}{c}{\textit{2.5k}} & \multicolumn{1}{c}{\textit{634}}\\
\midrule
\midrule
\multicolumn{15}{c}{Development Set Scores} \\
\midrule
\textbf{BERT} & 79.2 & 76.7 & 55.2 & 92.5 & 86.8 & 90.9 & \textbf{90.8} & \textbf{87.7} & 88.9 & 88.5 & 84.4 & 88.8 & 68.6 & 56.3 \\
\textbf{BERT$\rightarrow$QQP} & 78.6 & 76.0 & 49.7 & 91.5 & 84.3 & 89.0 & \sout{90.8} & \sout{87.7} & 89.7 & 89.5 & 83.7 & 87.7 & 72.6 & 56.3 \\
\textbf{BERT$\rightarrow$MNLI} & 81.1 & 79.2 & \textbf{59.0} & \textbf{92.7} & \textbf{88.5} & \textbf{91.9} & 90.8 & 87.5 & \textbf{90.3} & \textbf{90.2} & \sout{84.4} & \textbf{89.0} & \textbf{79.1} & 56.3 \\
\textbf{BERT$\rightarrow$SNLI} & 79.9 & 77.5 & 52.9 & \textbf{92.7} & 87.0 & 90.7 & 90.9 & 87.6 & 89.9 & 89.8 & \textbf{84.8} & 88.4 & 76.5 & 56.3 \\
\textbf{BERT$\rightarrow$Real/Fake} & 77.8 & 75.0 & 53.1 & 92.0 & 82.6 & 88.4 & 90.5 & 87.3 & 89.3 & 88.8 & 83.4 & 87.5 & 64.3 & 56.3 \\
\midrule
\textbf{BERT, Best of Each} & \textbf{81.2} & \textbf{79.3} & \textbf{59.0} & \textbf{92.7} & \textbf{88.5} & \textbf{91.9} & \textbf{90.8} & \textbf{87.7} & \textbf{90.3} & \textbf{90.2} & \textbf{84.8} & \textbf{89.0} & \textbf{79.1} & 56.3 \\
\midrule
\textbf{GPT} & 75.3 & 72.7 & \textbf{52.8} & \textbf{92.3} & 80.6 & 86.4 & 88.2 & 84.6 & 87.5 & 87.2 & 79.6 & 81.5 & 57.8 & 56.3 \\
\textbf{GPT$\rightarrow$QQP} & 73.1 & 69.7 & 29.8 & 91.4 & 82.8 & 87.7 & \sout{88.2} & \sout{84.6} & 87.4 & 87.3 & 80.1 & 78.9 & 62.8 & 56.3 \\
\textbf{GPT$\rightarrow$MNLI} & 76.2 & 74.1 & 41.5 & 91.9 & \textbf{86.8} & \textbf{90.8} & 88.8 & 81.3 & 89.2 & 89.0 & \sout{79.6} & \textbf{83.1} & \textbf{70.4} & 56.3 \\
\textbf{GPT$\rightarrow$SNLI} & 75.4 & 72.5 & 35.3 & 90.9 & 86.3 & 90.2 & \textbf{89.0} & \textbf{85.4} & \textbf{90.1} & \textbf{89.8} & \textbf{81.2} & 82.9 & 66.4 & 56.3 \\
\textbf{GPT$\rightarrow$Real/Fake} & 74.9 & 71.9 & 50.3 & 92.1 & 78.2 & 85.2 & 88.4 & 84.7 & 88.3 & 88.1 & \textbf{81.2} & 81.8 & 56.3 & 56.3 \\
\midrule
\textbf{GPT, Best of Each} & \textbf{78.0} & \textbf{75.9} & \textbf{52.8} & \textbf{92.3} & \textbf{86.8} & \textbf{90.8} & \textbf{89.0} & \textbf{85.4} & \textbf{90.1} & \textbf{89.8} & \textbf{81.2} & \textbf{83.1} & \textbf{70.4} & 56.3 \\
\bottomrule
\end{tabular}
\caption{Results on the GLUE development set with and without STILTs, fine-tuning on full training data of each target task. BERT results are based on the 12-layer model, while GPT results are \underline{without} an auxiliary language modeling objective. \textbf{Bold} indicates the best within each section. \sout{Strikethrough} indicates cases where the intermediate task is the same as the target task--we substitute the baseline result for that cell. \textit{A.Ex} is the average excluding MNLI and QQP because of the overlap with intermediate tasks. See text for discussion of WNLI results. }
\label{tab:glueresults_nolm}
\end{table*}

\end{document}